\pdfoutput=1

\documentclass[11pt]{article}

\usepackage[final]{acl}

\usepackage{times}
\usepackage{latexsym}

\usepackage[T1]{fontenc}

\usepackage[utf8]{inputenc}

\usepackage{microtype}

\usepackage{inconsolata}

\usepackage{graphicx}
\usepackage{multirow}
\usepackage{soul}
\usepackage{subcaption}
\usepackage{hyperref}

\usepackage{paralist}

\usepackage{color}

\title{MultiClaimNet: A Massively Multilingual Dataset of Fact-Checked Claim Clusters }

\author{Rrubaa Panchendrarajan\\
  Queen Mary University \\ of London \\
  \texttt{r.panchendrarajan} \\
  \texttt{@qmul.ac.uk} \\\And
  Rub\'{e}n M\'{i}guez\\
  Newtral Media Audiovisual \\
  Spain \\
  \texttt{ruben.miguez@newtral.es}\\\And
  Arkaitz Zubiaga \\
  Queen Mary University \\ of London \\
  \texttt{a.zubiaga@qmul.ac.uk}
  }

\begin{document}
\maketitle
\begin{abstract}
In the context of fact-checking, claims are often repeated across various platforms and in different languages, which can benefit from a process that reduces this redundancy. While retrieving previously fact-checked claims has been investigated as a solution, the growing number of unverified claims and expanding size of fact-checked databases calls for alternative, more efficient solutions. A promising solution is to group claims that discuss the same underlying facts into clusters to improve claim retrieval and validation. However, research on claim clustering is hindered by the lack of suitable datasets. To bridge this gap, we introduce \textit{MultiClaimNet}, a collection of three multilingual claim cluster datasets containing claims in 86 languages across diverse topics. Claim clusters are formed automatically from claim-matching pairs with limited manual intervention. We leverage two existing claim-matching datasets to form the smaller datasets within \textit{MultiClaimNet}. To build the larger dataset, we propose and validate an approach involving retrieval of approximate nearest neighbors to form candidate claim pairs and an automated annotation of claim similarity using large language models. This larger dataset contains 85.3K fact-checked claims written in 78 languages. We further conduct extensive experiments using various clustering techniques and sentence embedding models to establish baseline performance. Our datasets and findings provide a strong foundation for scalable claim clustering, contributing to efficient fact-checking pipelines.
\end{abstract}

\section{Introduction}
Automated fact-checking has become a crucial task to tackle the vast amount of unverified information circulating online. The core objectives of fact-checking pipelines are to identify claims that require verification, retrieve evidence, and assess their veracity automatically. The process can become even more challenging when the same claims are posted on different platforms in different languages \cite{smeros2021sciclops,quelle2023lost}. To overcome this challenge, the claim retrieval component in a fact-checking pipeline can retrieve, for each unverified claim, a set of previously fact-checked claim matches from a database, where available \cite{panchendrarajan2024claim}. 

As the number of verified and unverified claims grows, performing pairwise checks of each new claim against each of the database entries becomes inefficient and impractical for scalable fact-checking pipelines. Since claims are often repeated, an alternative solution is to form claim clusters by grouping the verified/unverified claims discussing the same underlying facts. This not only reduces redundancy in claim retrieval and validation but also enhances the efficiency and scalability of the fact-checking process. 

Research on identifying claim clusters has received limited attention in the literature, primarily due to the lack of suitable datasets. Existing studies have applied various clustering techniques to manually verify the existence of claim clusters \cite{kazemi-etal-2021-claim,nielsen2022mumin,quelle2023lost}. However, to the best of our knowledge no prior work has assessed the quality of the clusters due to the unavailability of datasets. Meanwhile, recent studies \cite{kazemi-etal-2021-claim,larraz2023semantic} have focused on annotating claim pairs that discuss the same underlying facts, enabling a more granular analysis of relationships between two claims. However, extending this to manual annotation of groups of claims that discuss the same fact is a more challenging and time-consuming task, which has hindered creation of datasets. 

In this research, we address this challenge by introducing and validating a methodology for data collection and labeling, and by automatically constructing multilingual claim cluster datasets from claim-matching pairs with limited manual intervention. To the best of our knowledge, this is the first work to create dedicated datasets for claim clustering. We present \textit{MultiClaimNet}, a collection of three multilingual claim cluster datasets.  The smaller datasets within \textit{MultiClaimNet} are derived from two existing claim-matching datasets, while the larger dataset is automatically constructed from the fact-checked claim dataset MultiClaim \cite{pikuliak2023multilingual}. Our approach for building this largest dataset involves retrieval of approximate nearest neighbors (ANN) from \textit{MultiClaim} to form candidate claim pairs, followed by automated similarity annotation using three large language models (LLMs). \textit{MultiClaimNet} comprises claims written in 86 unique languages across its three datasets, with the largest dataset containing 85.3K fact-checked claims. Furthermore, we conduct extensive experiments on the three datasets using various clustering approaches in combination with sentence embedding models, including LLMs, to establish baseline performance. We make the following key contributions:
\begin{compactitem}
    \item We present \textit{MultiClaimNet}\footnote{The dataset is available at \url{https://zenodo.org/uploads/15100352}}, a collection of three multilingual claim cluster datasets, constructed from claim-matching pairs with minimal manual intervention.
    \item We automatically generate the largest dataset in  \textit{MultiClaimNet} by leveraging a novel data collection and labeling methodology involving ANN retrieval and LLM annotation with no human intervention.
    \item We conduct extensive experiments on the three datasets with various clustering approaches and sentence embedding models to provide initial insights into the baseline performance. 
\end{compactitem}
We believe our datasets and findings will pave the way for further research in claim clustering, contributing to scalable solutions for automated fact-checking. 

\section{Related Work}
Claim clustering has been less widely studied than other subtasks within automated fact-checking, primarily due to the lack of available datasets. Most existing studies have only explored claim clustering as a means to validate the task and to check if clusters can be found.

\citet{kazemi-etal-2021-claim} introduced a multilingual claim matching dataset along with a classifier for detecting similar claims. Using this dataset, they applied a single-link hierarchical clustering, a variation of agglomeration clustering algorithm to confirm the existence of claim clusters. While their study identified meaningful multilingual clusters, it did not assess the quality of the clusters identified. \citet{hale2024analyzing} followed the same approach to perform an extensive manual analysis of claim clusters found in social media posts related to the Brazilian general election.

To address limitations arising from the unavailability of datasets, \citet{adler2019real} assumed that claims related to the same article belonged to the same cluster. They employed the DBSCAN clustering algorithm on the sentence embedding along with a community detection algorithm \cite{ester1996density} to merge smaller clusters. The authors performed a quantitative evaluation by measuring the fraction of claims that belong to the same news article.   

The most relevant study, conducted by \citet{quelle2023lost}, aimed to track claim evolution across languages. The authors represented claims as vectors using sentence embeddings and identified connected components by retrieving the most similar nearest neighbors. These connected components were considered claim clusters. However, none of the works created datasets with ground-truth labels, primarily due to the challenges associated with manual annotation.  

Instead of grouping claims discussing the same facts, several studies attempted to analyze the topics by grouping claims discussing the same topic. \citet{nielsen2022mumin} employed HDBSCAN \cite{mcinnes2017hdbscan} on sentence embeddings generated via sentence transformers \cite{reimers-2019-sentence-bert} to obtain more granular clusters discussing the same topic. \citet{smeros2021sciclops} used topic modeling techniques such as LDA \cite{lda} to generate topic vectors and then applied the KMeans algorithm to form the topic clusters. The authors used silhouette score \cite{ROUSSEEUW198753}, a widely used metric to evaluate unlabeled clusters. In contrast to these approaches, \citet{shliselberg2024syndy} leveraged large language models to annotate the topic associated with a claim and then trained a topic classifier in the synthetic dataset. while the topic labels can be used to form topic clusters, the number of distinct topics remains limited.

\section{Building the Claim Cluster Datasets}
We construct the claim cluster datasets from claim matching datasets, which consist of annotated claim pairs labeled as either similar or dissimilar based on whether they discuss the same underlying fact \cite{panchendrarajan2024claim}. We adhere to the definition of claim similarity provided by \citet{larraz2023semantic}. If two claims are labeled as similar, we assume the similarity is bidirectional; therefore, they belong to the same cluster. This assumption enables the creation of links between similar claim pairs, forming clusters of interconnected claims. For example, if Claim \textit{A} is annotated as similar to Claim \textit{B}, and Claim \textit{B} is similarly linked to Claim \textit{C}, we infer that Claims \textit{A}, \textit{B}, and \textit{C} belong to the same cluster. 

We utilized two existing multilingual claim matching datasets along with automatically annotated claim pairs using large language models to form three different claim cluster datasets. The following subsections detail the data sources and the methodology used to construct these multilingual claim cluster datasets. 

\begin{table*}[!t]
\footnotesize
\centering
\begin{tabular}{llllllll}\hline
Dataset       & Claim-pair Label & \# Claim Pairs                                                                      & \# Clusters & \# Claims & \begin{tabular}[c]{@{}l@{}}Avg. Cluster\\ Size\end{tabular} & \begin{tabular}[c]{@{}l@{}}Max \\ Cluster \\ Size\end{tabular} & \# Language  \\ \hline
ClaimCheck & \begin{tabular}[c]{@{}l@{}}Manual (Spanish)\\ Automated (rest)\end{tabular} & 5.2K & 197         & 1187      & 6.03                                                              & 28                                                             & 22                                                                              \\
ClaimMatch & Manual & 1.5K                                                                     & 192         & 1171      & 6.1                                                               & 35                                                             & 36                                                                            \\
MultiClaim & Automated &    54.4K                                                               & 30.9K       & 85.3K     & 2.76                                                              & 54                                                             & 78                              \\ \hline                                           
\end{tabular}
\caption{Statistics of \textit{MultiClaimNet}}\label{tab:data_stats}
\end{table*}

\subsection{Using Existing Claim Matching Datasets}

\subsubsection{Data Source}
We leveraged the following two claim matching datasets as the sources for creating the first two claim cluster datasets.

\noindent \textit{ClaimCheck} \cite{larraz2023semantic} - This dataset consists of 7.7K claim pairs collected using the Google FactCheck Explorer where each pair is annotated as either similar or dissimilar. The annotations were generated through a combination of manual and automated methods, with a 46-54 ratio. Automated annotation was performed using a similarity-based approach. The dataset primarily includes claims in Spanish and English, along with 20 other languages. For cluster creation, we utilized the 5.2K claim pairs labeled as similar.

\noindent \textit{ClaimMatch} - This dataset was obtained from the authors of \textit{ClaimCheck} \cite{larraz2023semantic} from Newtral Media Audiovisual, a fact-checking organization from Spain. This dataset is relatively smaller, with 2K claim pairs manually annotated. The authors used a set of queries or topics (e.g., Elon Musk) to curate these claims from the Google FactCheck API. For cluster creation, we utilized 1.5K similar pairs written in 36 languages. 

\subsubsection{Cluster Creation}
As mentioned earlier, we automatically generate clusters by linking claims shared among similar claim pairs. Using this approach, we clustered the claims from the two datasets discussed in the previous section, resulting in two multilingual claim clusters. However, this approach has the following two key limitations.
\begin{compactitem}
    \item Missing links - If two similar claims do not appear as similar pairs in the claim-matching dataset, this may not create the link between them, resulting in subclusters discussing the same fact.
    \item Merging of wrong clusters - If a statement discusses multiple claims, then creating links using such statements may result in the merging of claim clusters discussing different facts. 
\end{compactitem}

\begin{figure*}[!t]
    \centering
    \includegraphics[width=0.9\textwidth]{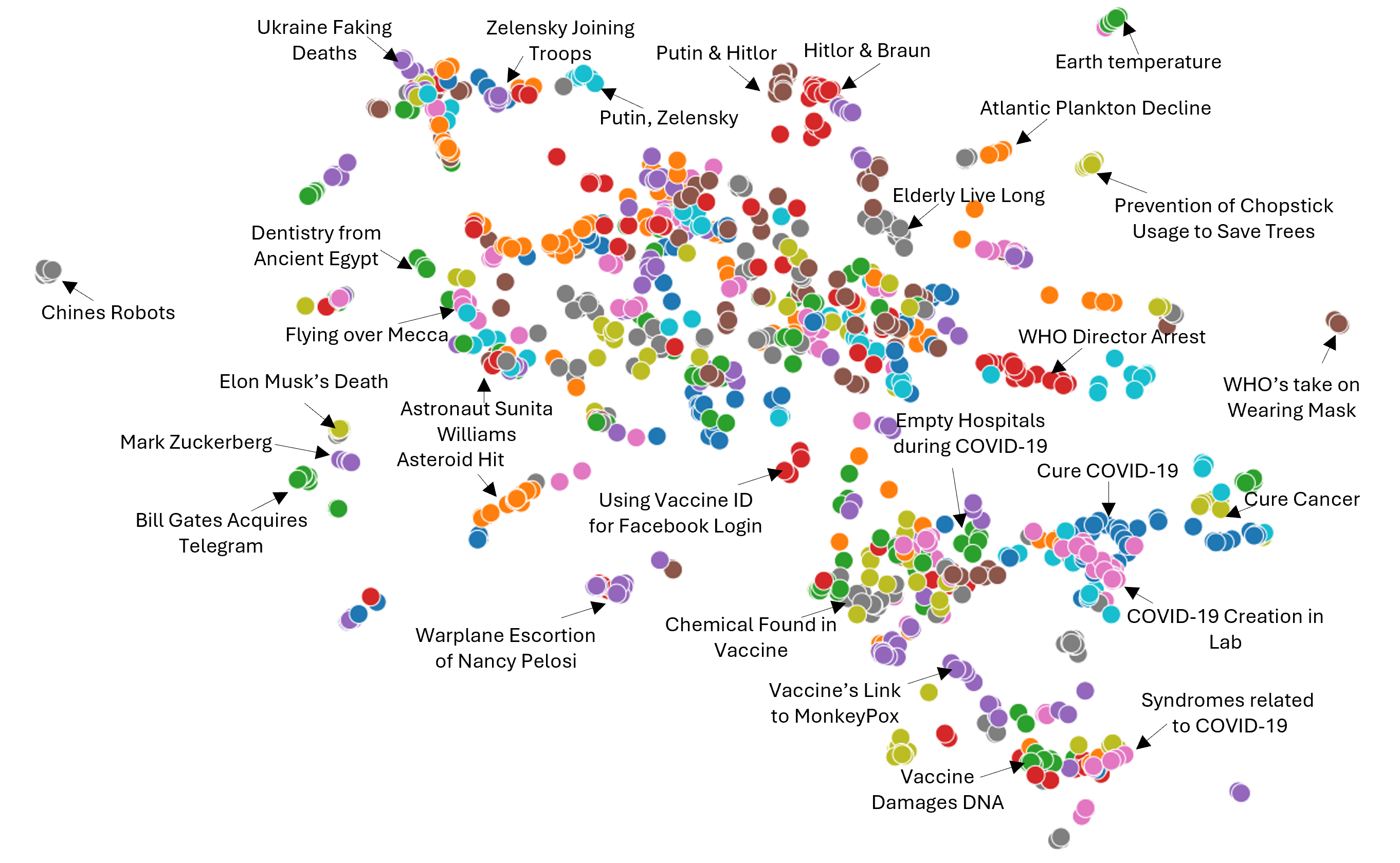}
    \caption{2D Visualization of ClaimCheck Clusters}
    \label{fig:claim_check_clusters}
\end{figure*}

To avoid the first limitation of missing links, we performed a manual inspection of clusters with high similarity. We converted the English translation of the claim to sentence embedding using Sentence Transformer \cite{reimers-2019-sentence-bert} and then computed the cluster embedding by averaging the sentence embedding of all claims within a cluster. \textit{ClaimMatch} dataset includes both original and translated claims. For the \textit{ClaimCheck} dataset, we translated the claims into English using Microsoft Azure AI translator \cite{junczys2019microsoft}. We then retrieved all the claim clusters with a cosine similarity greater than 0.75 and manually merged them into a single cluster if they discussed the same fact. This process resulted in only a small number of manual merges for both datasets.

To address the second limitation of merging wrong clusters, we manually validated clusters containing more than 20 claims for potential mismerges. However, we found that none of the clusters contained claims discussing different facts across both datasets, which further helped validate our approach.

Table \ref{tab:data_stats} presents the statistics of the two datasets. Despite having a higher number of claim pairs in \textit{ClaimCheck}, both datasets exhibit similar data patterns, except for the maximum cluster size and the number of languages present, both of which are slightly higher in \textit{ClaimMatch}. 

Figure \ref{fig:claim_check_clusters} presents the 2D visualization of the ground truth clusters of the \textit{ClaimCheck} dataset. 2D representations of claims were obtained using UMAP dimensionality reduction on the sentence embeddings of claims.  Claims within the same cluster are assigned the same color within a subspace (colors reused due to larger number of clusters). Some clusters in the figure are labeled with their respective topics. Interestingly, similar topics or concepts are placed closer together in the 2D space, and there exist smooth transitions in topics across the space. For instance, the progression from \textit{Vaccine} $\rightarrow$ \textit{COVID-19} $\rightarrow$ \textit{Wearing Mask} $\rightarrow$ \textit{WHO} $\rightarrow$ \textit{Environmental Issues} illustrates how clusters are interrelated. This suggests that beyond distinct claim clusters, broader conceptual groupings emerge, linking related topics across the space.

\begin{figure*}[!t]
    \centering
    \includegraphics[width=0.95\textwidth]{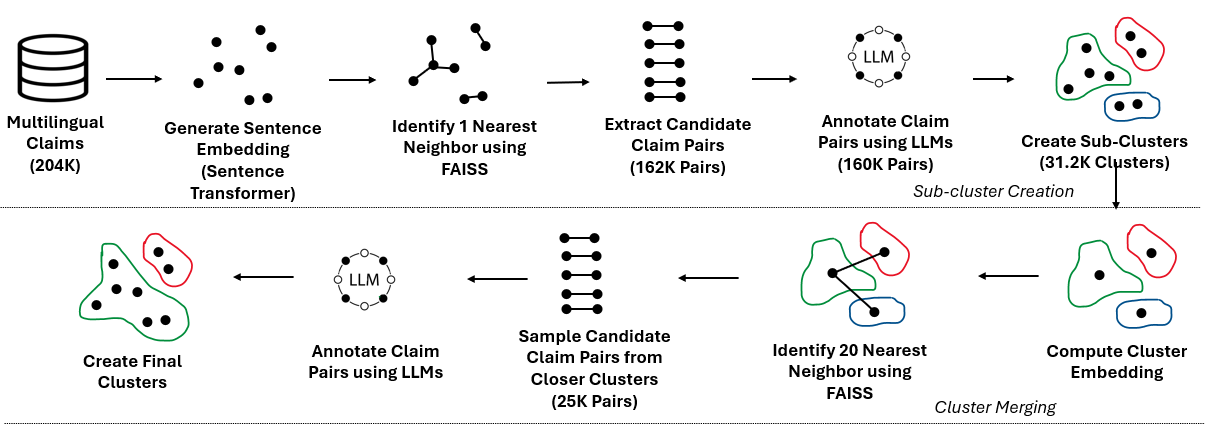}
    \caption{Methodology used for MultiClaim Dataset}
    \label{fig:methodology}
\end{figure*}

\subsection{MultiClaim Dataset}

The two datasets introduced earlier exhibit certain biases due to their curation process. Notably, \textit{ClaimCheck} contains similar pairs with higher semantic similarity, while \textit{ClaimMatch} covers only a limited set of topics. Further, both datasets are relatively small in size, which does not represent a real-world fact-checked database. To address these limitations, we automatically constructed a large-scale claim cluster dataset, mitigating biases related to similarity, topic coverage, and dataset size. 

For this study, we utilized the fact-checked multilingual claims from the MultiClaim \cite{pikuliak2023multilingual} dataset, which comprises 204K claims in 97 languages. The claims were primarily sourced from Google FactCheck Explorer. Given the dataset's extensive size, we propose a two-step approach for automatically constructing claim clusters. Figure \ref{fig:methodology} illustrates this approach, which is further detailed in the following sections.  

\subsubsection{Sub-Cluster Creation}
To identify groups of claims that discuss the same facts, we first need to detect similar claim pairs. Similar claims are likely to be positioned close together in a semantic vector space. Therefore, we employed an Approximate Nearest Neighbor (ANN) search to find the closest claim for each claim in the dataset. Specifically, we use Hierarchical Navigable Small Worlds (HNSW) \cite{malkov2018efficient}, one of the most widely used ANN algorithms. HNSW builds a multilayered graph and performs an approximate search for the nearest neighbor from the top to the bottom layer. Each claim is represented as an embedding vector of its English translation, generated using a Sentence Transformer. We then retrieve the nearest neighbor of each claim as a potential candidate for similar claim pairs. This resulted in 162K unique claim pairs (some claim pairs were duplicated because they were mutually the nearest neighbors of each other) to be annotated as similar or dissimilar. We retrieved only one nearest neighbor, as selecting more would significantly increase the volume of candidate pairs.

\begin{figure}[!t]
    \centering
    \includegraphics[width=0.42\textwidth]{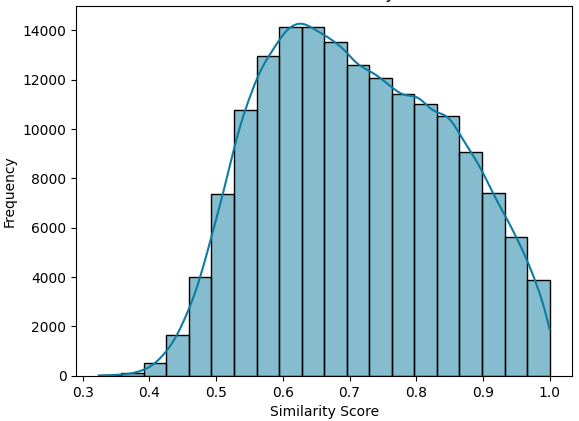}
    \caption{Similarity Distribution of Claim Pairs Extraction for Sub-cluster Creation}
    \label{fig:similarity distribution}
\end{figure}

Although we retrieved the closest neighbor for each claim, the similarity of the resulting candidate pairs ranged from 0.3 to 1, with an average of 0.71 (see Figure \ref{fig:similarity distribution}). Among the 162K claim pairs, some of them were an exact translation of the other claim, which we automatically annotated as similar. The remaining 160K claim pairs were annotated as similar or dissimilar using seven large language models (LLMs): Falcon 11B, Falcon 40B, GPT-4, Llama3 8B, Mistral 7B, Mixtral 8×7B, and Phi3 14B. We prompted these models with the question:

\textit{Do ‘Claim 1’ and ‘Claim 2’ discuss the same claim? Respond with Yes or No.}

\begin{figure}[!t]
    \centering
    \includegraphics[width=0.46\textwidth]{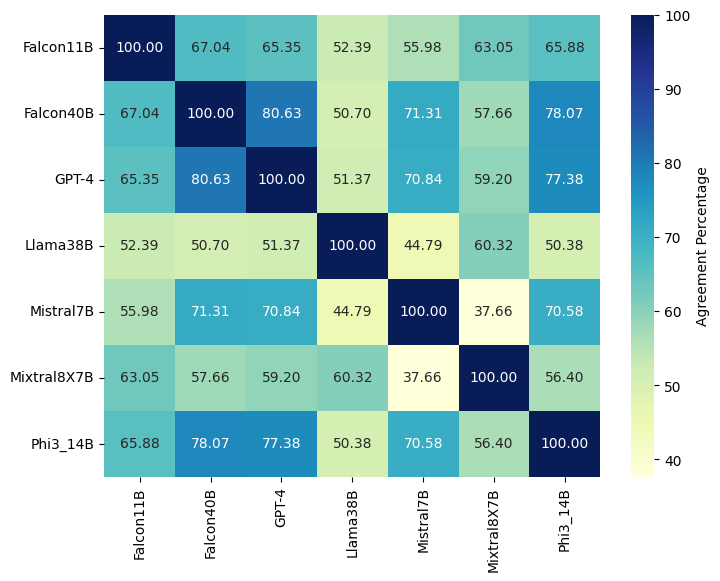}
    \caption{Percentage of Same Label Annotation among LLMs during Sub-cluster Creation}
    \label{fig:llm_aggrement}
\end{figure}

We deliberately used the term `same' instead of `similar' in the prompt, as the latter introduced more noise. Figure \ref{fig:llm_aggrement} shows the percentage of the same label annotated by the LLMs. Notably, GPT-4, Phi-3 14B, and Falcon 40B exhibited a higher agreement rate (77–80\%). To minimize noise in cluster creation, we selected only these three models as LLM annotators.

We used only the claim pairs that were unanimously annotated as similar by all three selected LLMs rather than relying on the majority label. While this approach may exclude some positive samples misclassified by one of the models, we found that relaxing the labeling criteria introduced more noise in subsequent steps. This process yielded 54K claim pairs consistently labeled as similar by all three LLMs. We followed a similar approach to construct claim clusters by linking these similar claim pairs. However, since we initially retrieved only one nearest neighbor per claim, this may lead to the splitting of clusters. Therefore, we refer to the generated clusters at this stage as sub-clusters and apply a merging process to refine them. In total, 31.2K sub-clusters were constructed at this stage.

\subsubsection{Cluster Merging}
We merge the sub-clusters generated in the previous step to eliminate biases, including the limitation of retrieving only a single nearest neighbor and the strict selection of claim pairs annotated as similar by all three LLMs. First, we represent each sub-cluster as an embedding by averaging the sentence embeddings of its claims. Next, we retrieve the top 20 nearest neighbors of each sub-cluster using HNSW and sample one claim per cluster to form candidate claim pairs. Since our clustering criteria is based on links between claim pairs, we believe that retrieving 20 nearest neighbors is sufficient to identify missing links between sub-clusters. Similar to the manual validation, we filtered out claim pairs from clusters with a cosine similarity of less than 0.75. This process yields 25K claim pairs, which are then annotated as similar or dissimilar by LLMs. 

During the annotation of inter-cluster claim pairs, we observed a decrease in label agreement across LLMs, ranging from 41\% to 69\%. This suggests that annotating inter-cluster claim pairs was more challenging for the LLMs compared to the annotation of nearest neighbors. However, we only considered the 8.5K claim pairs annotated with the same label by all three LLMs as valid, as relaxing this criterion led to mismerges. This process yielded a final set of 30.9K claim clusters. We refer to this dataset as \textit{MultiClaim}, with its statistics detailed in Table \ref{tab:data_stats}. 


\subsection{Multilingual Claims}
Our \textit{MultiClaimNet} dataset encompasses multilingual claims written in 86 unique languages across the three cluster datasets. Refer to Appendix \ref{appendix:language_stats} for the language-wise statistics of the datasets. The smaller datasets exhibit similar language distributions, with Spanish being the dominant language in the \textit{ClaimCheck} dataset, and English as the majority language in \textit{ClaimMatch}. Additionally, both of these datasets predominantly feature European languages. In contrast, \textit{MultiClaim} includes a broader mix of both Asian and European languages, with English remaining the dominant language.

\begin{table}[!t]
\footnotesize
\begin{tabular}{lcc}\hline
Dataset       &  \begin{tabular}[c]{@{}c@{}}Monolingual vs \\ Multilingual \\ Clusters\end{tabular} & \begin{tabular}[c]{@{}c@{}}Avg. Unique \\ Languages in \\ Multilingual Clusters\end{tabular}\\ \hline
ClaimCheck & 55/142 & 3.2                                                                             \\
ClaimMatch & 58/134 & 3.8                                                                            \\
MultiClaim & 15.9K/15K   & 2.4                           \\ \hline                                           
\end{tabular}
\caption{Statistics of Multilingual Clusters}\label{tab:language_stats}
\end{table}

Table \ref{tab:language_stats} presents the statistics of multilingual clusters present in the cluster datasets. Notably, the smaller datasets contain a higher proportion of multilingual clusters, whereas the larger one has an equal number of monolingual and multilingual clusters. Furthermore, the multilingual claim clusters consist of claims written in 2.4-3.8 unique languages on average, validating the existence of similar claims across different languages. Further, this highlights the importance of advancing multilingual research in automated fact-checking to effectively handle repeated claims across diverse linguistic contexts.

\begin{figure}[!t]
    \centering
    \includegraphics[width=0.48\textwidth]{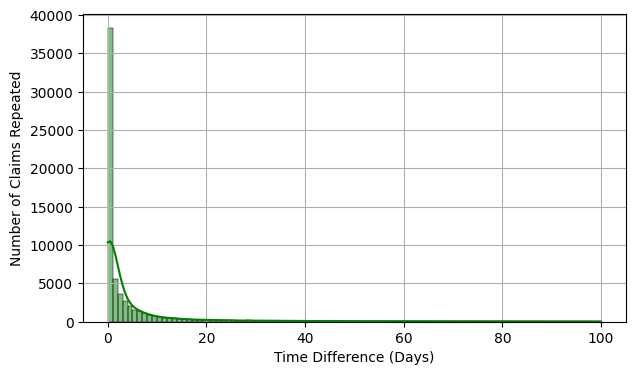}
    \caption{Number of Claims Repeated over First 100 Days in \textit{MultiClaim} Clusters}
    \label{fig:claim_repition}
\end{figure}

\subsection{Temporal Claims}
The reoccurrence of the same claim demands techniques to tackle repeated claims for an effective fact-checking pipeline. Among the three datasets, \textit{MultiClaim} contains the timestamps of the claims, making it an ideal choice for further research on temporal claims. Figure \ref{fig:claim_repition} illustrates the distribution of claim repetitions over the first 100 days. Notably, 50\% of the claims are repeated within just 1.6 days, and 75\% are repeated within the first 29.3 days from the occurrence of the initial claim in a cluster.  This underscores the need for advanced methods to detect and manage repeated claims efficiently, ensuring that fact-checking systems can respond swiftly and accurately to the rapid spread of misinformation.

\section{Experiment Setup}
We employ various clustering techniques to evaluate the performance of baseline models on the claim cluster datasets. 

\subsection{Clustering Approaches}
In real-world fact-checked databases, the number of clusters is often unknown. Therefore, we utilize clustering techniques that do not require the number of clusters as a predefined parameter. Instead, these methods automatically determine the optimal number of clusters based on other controlling parameters, such as density thresholds or distance metrics. We apply the following clustering methods to the sentence vectors generated by sentence embedding models.
\begin{compactitem}
    \item HDBSCAN \cite{mcinnes2017hdbscan} - A hierarchical density-based clustering algorithm that determines dense regions and merges them to form hierarchical trees.
    \item Agglomerative clustering \cite{mullner2011modern} - A hierarchical clustering method that builds tree-like structures and iteratively merges them using links.
    \item Affinity Propagation \cite{dueck2009affinity} - A message-passing clustering algorithm that identifies representative points and generates clusters by assigning other points to them based on similarity. 
    \item Birch \cite{zhang1996birch} - A scalable algorithm designed for large datasets to form tree structures and merge trees iteratively. 
    \item MeanShift \cite{comaniciu2002mean} - A centroid-based clustering algorithm that iteratively shifts points toward high-density regions to form clusters automatically
    \item Optics \cite{ankerst1999optics} -  A density-based clustering algorithm that identifies dense regions from points ordered according to their reachability distance.  
\end{compactitem}

The hyperparameters used for the clustering algorithms and parameter search are discussed in Appendix \ref{appendix:hyperparameters}. Density-based approaches are shown to be very effective when combined with dimensionality reduction techniques such as UMAP \cite{nielsen2022mumin}. Therefore, we reduce the sentence embedding into 8 dimensions (optimal dimension across all three datasets) for the density-based approaches HDBSCAN and Optics.

\begin{table}[]
\centering
\footnotesize
\begin{tabular}{p{5cm}ll} \hline
Model                                          & \rotatebox{90}{ Parameters } & \rotatebox{90}{ Embedding } \\ \hline
Distiluse-base-multilingual-cased-v1           & 135M & 512       \\
Paraphrase-multilingual-MiniLM-L12-v2          & 118M & 768       \\
Paraphrase-multilingual-mpnet-base-v2          & 278M & 768       \\
Gte-multilingual-base                          & 305M & 768      \\
All-roberta-large-v1                           & 355M & 1024      \\
LaBSE                                          & 471M & 768       \\
KaLM-embedding-multilingual-mini-instruct-v1.5 & 494M & 896       \\
Multilingual-e5-large-instruct                 & 560M & 1024      \\
Bge-m3                                         & 567M & 1024      \\ \hline
Gte-Qwen2-1.5B-instruct                        & 1B   & 1536      \\
MiniCPM-Embedding                              & 2.4B & 2304      \\
E5-mistral-7b-instruct                         & 7B   & 4096      \\
Gte-Qwen2-7B-instruct                          & 7B   & 3584      \\
LLM2Vec-Meta-Llama-3-8B-Instruct-mntp          & 8B   & 4096      \\
Bge-multilingual-gemma2                        & 9B   & 3584    \\ \hline 
\end{tabular}
\caption{Multilingual Sentence Embedding Models}\label{tab:sentence_embedding}
\end{table}

\subsection{Multilingual Models}
Unlike dataset curation, where we used the English translation, we explore multilingual representations of claims by encoding their original text as sentence embeddings. A widely used approach for generating sentence embeddings is Sentence Transformers \cite{reimers-2019-sentence-bert}, which offers pretrained models optimized for semantic similarity tasks. In our experiments, we evaluate sixteen multilingual sentence embedding models listed in Table \ref{tab:sentence_embedding}. All models, except \textit{LLM2Vec-Meta-Llama-3-8B-Instruct-mntp} \cite{llm2vec}, are available as Sentence Transformers. \textit{LLM2Vec-Meta-Llama-3-8B-Instruct-mntp} is a text encoder model derived from the Llama 3-8B large language model, designed specifically for generating text representations. 

\begin{table*}[!t]
\centering
\footnotesize
\begin{tabular}{lllllllll} \hline
Dataset                        & Approach            & \# Clusters & ARI   & AMI   & HMG & CMP & V-Measure & Purity \\ \hline
\multirow{6}{*}{ClaimCheck} & HDBScan             & 188            & 0.794 & 0.89  & 0.952       & 0.96         & 0.956     & 0.893  \\
                               & Agglomerative       & 275            & 0.723 & 0.874 & \textbf{0.994}       & 0.922        & 0.956     & \textbf{0.987}  \\
                               & AffinityPropagation & 183            & \textbf{0.806} & \textbf{0.9}   & 0.956       & \textbf{0.963}        & \textbf{0.96}      & 0.919  \\
                               & Birch               & 196            & 0.551 & 0.827 & 0.894       & 0.957        & 0.925     & 0.796  \\
                               & MeanShift           & 212            & 0.116 & 0.629 & 0.743       & 0.916        & 0.821     & 0.637  \\
                               & Optics              & 258            & 0.051 & 0.53  & 0.749       & 0.857        & 0.799     & 0.702  \\ \hline
\multirow{6}{*}{ClaimMatch} & HDBScan             & 191            & 0.584 & 0.779 & 0.908       & 0.913        & 0.91      & 0.819  \\
                               & Agglomerative       & 292            & 0.591 & 0.791 & \textbf{0.978}       & 0.885        & \textbf{0.93}      & \textbf{0.946}  \\
                               & AffinityPropagation & 158            & \textbf{0.68}  & \textbf{0.819} & 0.915       & \textbf{0.934}        & 0.924     & 0.833  \\
                               & Birch               & 99             & 0.269 & 0.64  & 0.688       & 0.931        & 0.792     & 0.437  \\
                               & MeanShift           & 186            & 0.041 & 0.413 & 0.588       & 0.83         & 0.688     & 0.439  \\
                               & Optics              & 284            & 0.11  & 0.537 & 0.828       & 0.821        & 0.825     & 0.759  \\ \hline
\multirow{4}{*}{MultiClaim} & HDBScan             & 10599          & 0.007 & 0.383 & 0.733       & \textbf{0.975}        & 0.837     & 0.339  \\
                               & Agglomerative       & 27853          & \textbf{0.574} & \textbf{0.714} & \textbf{0.961}       & 0.973        & \textbf{0.967}     & \textbf{0.784}  \\
                               & Birch               & 32034          & 0.071 & 0.398 & 0.873       & 0.946        & 0.908     & 0.605  \\
                               & Optics              & 21509          & 0.001 & 0.221 & 0.71        & 0.935        & 0.807     & 0.519 \\ \hline
\end{tabular}
\caption{Performance of Different Clustering Approaches}\label{tab:performance_clustering_methods}
\end{table*}

\subsection{Metrics}
We report the following metrics, which measure different aspects of the clusters generated against the ground-truth clusters \cite{pauletic2019overview}.

\begin{compactitem}
    \item Adjusted Rand Index (ARI) - Measures similarity between two clusters
    \item Adjusted Mutual Index (AMI) - Measures mutual information shared between two clusters
    \item Homogeneity (HMG) - Measures the fraction of cluster instances belonging to the same ground-truth cluster
    \item Completeness (CMP) - Measures the fraction of ground-truth cluster instances that are grouped together
    \item V-Measure (VM) - Measures the harmonic mean of homogeneity and completeness
    \item Purity - Measures chances of cluster instances belonging to the same ground truth cluster
\end{compactitem}

\section{Results}

\subsection{Clustering Approaches}

Table \ref{tab:performance_clustering_methods} presents the performance of different clustering approaches across the three datasets. For this experiment, we use the \textit{Bge-m3} \cite{bge-m3} model as the sentence transformer, as it performed well across all datasets. Results for the Affinity propagation and MeanShift approaches are not reported for \textit{MultiClaim} due to high memory and running time requirements. However, Affinity propagation generally performs well across the smaller dataset, achieving the highest scores across multiple metrics. 

Agglomerative clustering is the only approach that consistently 
performed well in the largest dataset across all metrics. This highlights its potential for integration in large-scale fact-check databases. However, its performance slightly declines on smaller datasets as they tend to generate more fine-grained clusters (i.e., a larger number of clusters than the ground truth), leading to higher homogeneity and purity scores. Parameter tuning revealed that Agglomerative clustering could achieve better performance on smaller datasets if different parameter sets were used based on dataset size. However, for a fair comparison, we use the same parameters across all three datasets, which yielded the near-optimal. 

Another limitation of Agglomerative clustering is its significant increase in runtime as the number of claims grows. HDBSCAN efficiently clusters both small and large datasets. However, in \textit{MultiClaim}, it tends to perform poorly, as the reduction in dimensionality consistently leads to excessively granular clusters. In other words, HDBSCAN struggles to find fine-grained clusters when the number of data points is increased. 

\begin{table*}[!t]
\centering
\footnotesize
\begin{tabular}{p{5cm}lllllllll} \hline
                                               & \multicolumn{3}{c}{ClaimCheck}                & \multicolumn{3}{c}{ClaimMatch}                & \multicolumn{3}{c}{MultiClaim}                \\ \hline
                                               & ARI            & AMI            & VM      & ARI            & AMI            & VM      & ARI            & AMI            & VM      \\ \hline
distiluse-base-multilingual-cased-v1           & 0.544          & 0.752          & 0.924          & 0.425          & 0.635          & 0.894          & 0.222          & 0.477          & 0.939          \\
paraphrase-multilingual-MiniLM-L12-v2          & 0.657          & 0.821          & 0.941          & 0.552          & 0.743          & 0.915          & 0.331          & 0.564          & 0.947          \\
paraphrase-multilingual-mpnet-base-v2          & 0.702          & 0.854          & 0.95           & 0.606          & 0.788          & 0.927          & 0.488          & 0.639          & 0.957          \\
gte-multilingual-base                          & \textbf{0.845} & \textbf{0.919} & \textbf{0.969} & \textbf{0.711} & \textbf{0.842} & \textbf{0.941} & 0.467          & 0.656          & 0.952          \\
all-roberta-large-v1                           & 0.411          & 0.634          & 0.892          & 0.328          & 0.522          & 0.855          & 0.111          & 0.344          & 0.908          \\
LaBSE                                          & 0.586          & 0.772          & 0.928          & 0.462          & 0.672          & 0.899          & 0.506          & 0.631          & 0.96           \\
KaLM-embedding-multilingual-mini-instruct-v1.5 & 0.703          & 0.826          & 0.925          & 0.489          & 0.686          & 0.86           & 0.138          & 0.371          & 0.872          \\
multilingual-e5-large-instruct                 & 0.732          & 0.85           & 0.931          & 0.545          & 0.738          & 0.875          & 0.132          & 0.391          & 0.867          \\
bge-m3                                         & 0.723          & 0.874          & 0.956          & 0.591          & 0.791          & 0.93           & \textbf{0.574} & \textbf{0.714} & \textbf{0.967} \\ \hline
gte-Qwen2-1.5B-instruct                        & 0.587          & 0.796          & 0.935          & 0.478          & 0.708          & 0.909          & 0.443          & 0.569          & 0.952          \\
MiniCPM-Embedding                              & 0.585          & 0.783          & 0.931          & 0.439          & 0.654          & 0.893          & 0.301          & 0.489          & 0.938          \\
gte-Qwen2-7B-instruct                          & 0.665          & 0.835          & 0.945          & 0.51           & 0.732          & 0.915          & 0.493          & 0.634          & 0.958          \\
e5-mistral-7b-instruct                         & 0.821          & 0.905          & 0.963          & 0.626          & 0.789          & 0.921          & 0.268          & 0.465          & 0.914          \\
LLM2Vec-Meta-Llama-3-8B-Instruct-mntp          & 0.285          & 0.47           & 0.786          & 0.208          & 0.372          & 0.744          & 0.045          & 0.219          & 0.836          \\
bge-multilingual-gemma2                        & 0.607          & 0.798          & 0.933          & 0.507          & 0.73           & 0.913          & 0.438          & 0.591          & 0.952     \\ \hline    
\end{tabular}
\caption{Performance of Different Sentence Embedding Models}\label{tab:performance_models}
\end{table*}

\subsection{Sentence Embedding Models}

Table \ref{tab:performance_models} presents the performance of various sentence embedding models when applied with Agglomerative clustering. Interestingly, smaller models (< 1B parameters) achieved the highest scores compared to the larger models. In particular, \textit{gte-multilingual-base} \cite{zhang2024mgte} achieves the highest performance across small datasets, while \textit{Bge-m3} \cite{bge-m3} outperforms others on the \textit{MultiClaim} dataset. This suggests that large language models are not necessarily required for effective sentence representation in semantic similarity tasks such as clustering. Moreover, the multilingual capabilities of these models could facilitate further research by mitigating language barriers. 

However, we observed that clustering algorithms are highly sensitive to sentence representation unless combined with a dimensional reduction technique. For instance, Table \ref{tab:avg_clusters} presents the average number of clusters and their corresponding standard deviation across different sentence embedding models combined with HDBSCAN and Agglomerative clustering. Agglomerative exhibits greater sensitivity to the sentence representation, leading to a 4-7.5 fold increase in standard deviation compared to HDBSCAN. This suggests that developing more robust clustering solutions, independent of the number of data instances and sentence representations, is essential for future research.

\begin{table}[]
\footnotesize
\begin{tabular}{lll} \hline
              & HDBSACN & Agglomerative \\ \hline
ClaimCheck & 207 $\pm$ 19    & 276  $\pm$ 92         \\
ClaimMatch & 215 $\pm$ 14   & 290  $\pm$  104       \\
MultiClaim & 8.5K $\pm$  2.3K & 22.1K $\pm$ 9.4K      \\ \hline
\end{tabular}
\caption{Average Number of Clusters Found using Different Sentence Embedding Models}\label{tab:avg_clusters}
\end{table}

\begin{table*}[]
\footnotesize
\begin{tabular}{lp{11.3cm}} \hline
Scenario                                                                                                                   & Misgrouped Claims (English Translation)                                                                                                                                                                          \\ \hline
\multirow{2}{*}{\begin{tabular}[c]{@{}l@{}}Not recognizing the same  \\entities with different reference\end{tabular}} & AstraZeneca vaccine against covid-19 causes monkeypox                                                                                                                            \\
                                                                                                                       & Monkeypox is a consequence of anti-covid vaccination                                                                                                                             \\ \hline
\multirow{2}{*}{\begin{tabular}[c]{@{}l@{}}Not knowing the background \\ of the entities\end{tabular}}                 & Tedros Adhanom was arrested by Interpol                                                                                                                                          \\
                                                                                                                       & WHO Chief Arrested For Crimes Against Humanity                                                                                                                                   \\ \hline
\multirow{2}{*}{Not focusing on keywords}                                                                              & The American Rescue Plan helped create nearly 10 million new jobs.                                                                                                               \\
                                                                                                                       & Joe Biden states that the American Rescue Plan helped create nearly 10 million new jobs.                                                                                         \\ \hline
\multirow{2}{*}{Prone to noise}                                                                                        & A satirical article virally shared via social media 'reported' that World Health Organization director Tedros Adhanom Ghebreyesus had been arrested for crimes against humanity. \\
                                                                                                                       & Tedros Adhanom was arrested by Interpol                                                                                                                                          \\ \hline
\multirow{2}{*}{Not recognizing implicit claim}                                                                        & Picture shows Zelenskyy's Russian passport.                                                                                                                                      \\
                                                                                                                       & Zelensky has a Russian passport and not a Ukrainian passport                                                                                                                     \\ \hline
\multirow{2}{*}{Merge due to common entities}                                                                       & Maduro has threatened to send missiles to Spain                                                                                                                                  \\
                                                                                                                       & Maduro asks for support for Brazil's military to overthrow Bolsonaro                                                                                           \\ \hline                 
\end{tabular}
\caption{Error Scenarios and Sample Misgrouped Claims}\label{tab:errors}
\end{table*}

\subsection{Error Analysis}
We conduct an error analysis of the clusters formed using the baseline methods. Specifically, we apply HDBSCAN with \textit{Gte-multilingual-base} embeddings for the smaller datasets and  Agglomerative clustering using \textit{Bge-m3} embeddings on the larger datasets. Table \ref{tab:errors} lists some of the error scenarios. Except for the last, \textit{merge due to common entities}, other scenarios result in misplacing claims discussing the same facts in different clusters. This finding highlights the limitations of the baseline approaches and suggests the need for task-specific solutions for obtaining accurate claim clusters. 

\section{Applications}
Although claim clustering can be integrated across various components of automated fact-checking, we outline key applications of our datasets. 
\begin{compactitem}
    \item Claim cluster database - Verified claims can be maintained as claim clusters by grouping the claims discussing the same facts. Traditional clustering techniques can be further explored to group multilingual claims at a larger scale. 
    \item Fact-checked claim cluster retrieval - Instead of retrieving individual verified claims from a fact-checked dataset, an unverified claim can be matched against an entire cluster, offering a broader context of the claim across different occurrences of the claim in different languages. 
    \item Iterative claim clustering - Clustering approaches can be explored further to perform iterative clustering representing real-world scenarios of integration of new claims, where it can be merged into an existing cluster or form a new cluster.
    \item Verification of claim clusters - Similar to the grouping of verified claims, unverified claims can be clustered together, streamlining the verification process by allowing entire clusters to be assessed collectively.
    \item Visualize topic themes - Claim clusters can be further grouped using techniques such as hierarchical clustering to analyze inter-related claim clusters and to identify topic themes.
\end{compactitem}

\section{Conclusion}
This paper introduces \textit{MultiClaimNet}, a collection of three multilingual claim clustering datasets constructed from similar claim pairs. The largest dataset within \textit{MultiClaimNet} was created by retrieving claim pairs using approximate nearest neighbor approaches and annotating them with large language models. We conduct extensive experiments on these datasets, evaluating different clustering approaches and multilingual sentence embedding models. Our results show that Agglomeration clustering performs well on the largest dataset and remains competitive across the smaller ones. In contrast, HDBSCAN excels only on the smaller datasets, underscoring the need for robust solutions. Additionally, the smaller sentence embedding models outperform larger ones, highlighting their potential for scalable fact-checking solutions. Our error analysis further suggests that task-specific approaches are essential for improving clustering accuracy. We further highlight the potential applications of our dataset and believe it will be a valuable resource for advancing multilingual fact-checking research.

\section*{Limitations}\label{sec:limitation}
The limitations of this work are as follows:
\begin{compactitem}
    \item Data curation bias - As discussed earlier, the smaller dataset introduces biases such as highly similar claim pairs and a limited range of topics due to the data curation techniques.
    \item Reliance on textual content alone for annotation - Claim pairs in \textit{MultiClaim} dataset were annotated for their similarity based solely on textual content presented to LLMs. This may result in false positives when other media associated with a claim refer to different incidents. 
    \item Mismerge due to text with multiple claims - We assume that a fact-checked claim generally discusses one factual statement. However, the presence of claims with more than one factual statement may create false positive links between claim pairs. 
\end{compactitem}

\section*{Acknowledgments}
This project is funded by the European Union and UK Research and Innovation under Grant No. 101073351 as part of Marie Skłodowska-Curie Actions (MSCA Hybrid Intelligence to monitor, promote, and analyze transformations in good democracy practices). We gratefully acknowledge Neutral Media Audiovisual for providing the \textit{ClaimCheck} and \textit{ClaimMatch} claim-matching datasets. We acknowledge Queen Mary's Apocrita HPC facility, supported by QMUL Research-IT, for enabling our experiments \cite{king_2017_438045}.

\bibliography{references}

\appendix

\newpage

\section{Appendix}

\subsection{Sample Claims}
Table \ref{tab:sample_claims} presents sample claims from the largest cluster in each dataset. Claims about the same fact vary in length and tend to refer to the same entities related to the claim with various references. As mentioned in Limitations (Section \ref{sec:limitation}), the LLM-based annotation for claim similarity may introduce false positive samples when the media associated with the claims refer to different incidents. For example, in \textit{MultiClaim}, the largest cluster consists of claims about Russian attacks in Ukraine. While these claims share similar textual descriptions, they may refer to different events within the war.

\begin{table*}[!th]
\centering
\small
\begin{tabular}{lp{13cm}} \hline
Dataset                        & Factchecked Claim (English Translation)                                                                                                                      \\ \hline
\multirow{3}{*}{ClaimCheck} & COVID-19 vaccines contain graphene oxide                                                                                                      \\
                               & Former Pfizer employee confirms vaccine contains graphene oxide nanoparticles                                                                 \\
                               & A Facebook post claimed that Spanish doctors had detected graphene oxide in Covid vaccines                                                    \\ \hline
\multirow{3}{*}{ClaimMatch} & This video proves chemtrail                                                                                                                   \\
                               & Harmful substances are being dropped from airplanes.                                                                                          \\
                               & The claim that the video shows the moment when the pilot mistakenly released harmful chemicals referred to as chemtrails at the airport       \\ \hline
\multirow{3}{*}{MultiClaim} & Video shows Russian forces attacking Ukraine                                                                                                  \\
                               & A video of a recent Russian airstrike on Ukraine                                                                                              \\
                               & The video shows a large explosion that took place in Ukraine after Russian President Vladimir Putin announced a military operation in Ukraine  \\ \hline
\end{tabular}
\caption{Sample FactChecked Claims from the Largest Clusters}\label{tab:sample_claims}
\end{table*}

\subsection{Language Statistics}\label{appendix:language_stats}
\begin{figure*}[!th]
    \centering
    \begin{subfigure}[b]{0.75\textwidth}
    \includegraphics[width=\textwidth]{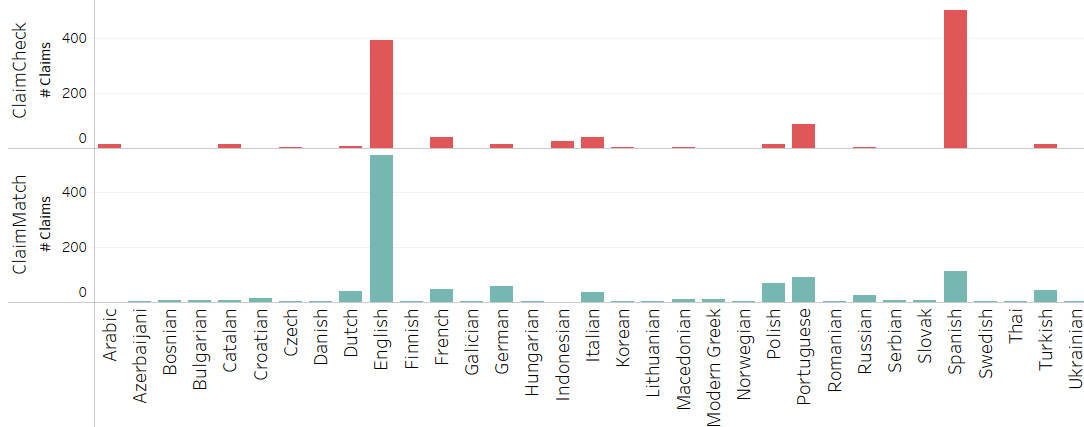}
    \caption{ClaimCheck and ClaimMatch}
    \label{fig:language_stats_cc_cm}   
    \end{subfigure}  
    \begin{subfigure}[b]{1\textwidth}
    \includegraphics[width=\textwidth]{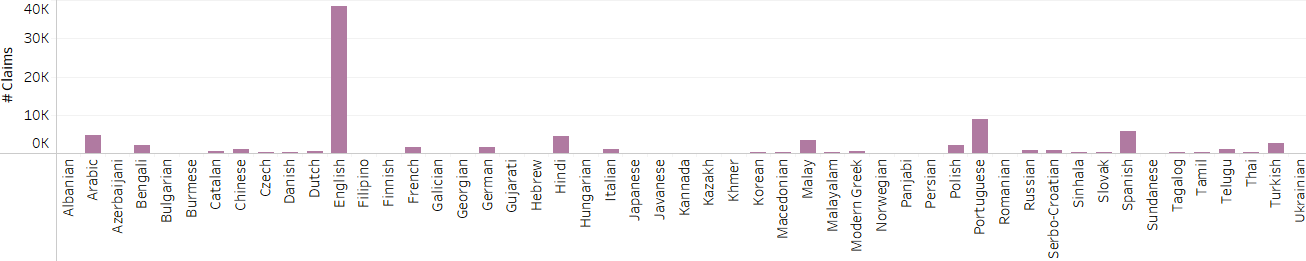}
    \caption{MultiClaim}
    \label{fig:language_stats_mc}   
    \end{subfigure}
    \caption{Statistics of Language Occurrence in the \textit{MultiClaimNet} Dataset}
    \label{fig:language_stats}
\end{figure*}

Figure \ref{fig:language_stats} presents the language statistics across three datasets, including only languages that appear at least three times in a dataset.

\subsection{Hyperparameter Tuning}
\label{appendix:hyperparameters}

\begin{table*}[!ht]
\small
\centering
\begin{tabular}{llll}
\hline
Approach                       & Hyperparameter            & Search Options                        & Optimal \\ \hline
\multirow{4}{*}{HDBSAN}        & vector dimenstion         & {[}2,3,4,8,16,32,64,128,256,512{]}    & 8       \\ 
                               & cluster selection epsilon & Minimum value                         & 0.1     \\ 
                               & min samples               & Minimum value                         & 1       \\ 
                               & min cluster size          & Minimum value                         & 2       \\ \hline
\multirow{2}{*}{Agglomerative} & distance threshold         & {[}0.5 - 2{]}, step size of 0.5       & 1       \\ 
                               & linkage                   & {[}ward, complete, average, single{]} & ward    \\ \hline
Birch                          & threshold                 & {[}0.1 - 1{]}, step size 0.1          & 0.7     \\ \hline
Meanshift                      & bandwidth                 & {[}0.1 - 0.8{]}, step size of 0.05    & 0.75    \\ \hline
OPTICS                         & min samples               & Minimum value                         & 2       \\ \hline
\end{tabular}
\caption{Hyperparameters of Clustering Approaches}\label{tab:hyperparameters}
\end{table*}

We use the Scikit-learn\footnote{https://scikit-learn.org/stable/api/sklearn.cluster.html} library for the execution of clustering algorithms. The optimal hyperparameters that yield the best performance in terms of V-Measure across all three datasets were used to report the results. Table \ref{appendix:hyperparameters} lists the hyperparameter search and the optimal settings used. Apart from these hyperparameters, we use the default parameters recommended in the library for each clustering approach.

\end{document}